\documentclass[runningheads]{llncs}

\usepackage[mobile]{eccv}

\usepackage{eccvabbrv}

\usepackage{placeins}

\usepackage{graphicx}
\usepackage{booktabs}

\usepackage[accsupp]{axessibility}  

\usepackage{hyperref}  

\usepackage{orcidlink}

\begin{document}

\title{Media Intelligence Platform: Multimodal Detection of Synthetic and Manipulated Content} 

\titlerunning{Abbreviated paper title}

\author{David Fischinger, \inst{1}\orcidlink{0000-0002-0806-0468} \and
Marcel Hasenbalg \inst{1}\orcidlink{0000-0001-8012-8669} \and
Silvia Poletti\inst{1}\orcidlink{0000-0002-5923-531X} \and
Davide Antonutti \inst{1} \orcidlink{0009-0006-0454-2669} \and
Mina Schütz \inst{1} \orcidlink{0000-0002-1102-306X} \and
Lam Pham \inst{1} \orcidlink{0000-0001-8155-7553} \and
Justin Ilyes \inst{1} \orcidlink{0009-0000-6343-6176} \and
Péter Rjabcsenkó \inst{1} \orcidlink{0009-0004-3868-9779} \and
Simon Freitter \inst{} \and
Martin Boyer \inst{1} \orcidlink{0000-0002-0765-0829}}

\authorrunning{D.~Fischinger et al.}

\institute{Austrian Institute of Technology, Giefinggasse 4, 1210 Vienna - Austria
\email{forname.surname@ait.ac.at}\\
\url{https://www.ait.ac.at/en/}}

\maketitle

\begin{abstract}
The rapid progress of generative artificial intelligence has led to a significant increase in manipulated multimedia content, including synthetic faces, forged videos, AI-generated images, fake audio, and locally edited images. Existing detection approaches are typically designed for a single manipulation type, limiting their applicability in real-world scenarios involving heterogeneous media sources. In this work, we present the Media Intelligence Platform (MIP), a unified web-based framework for multimodal media forensics integrating specialized modules for fake video detection, face morphing detection, AI-generated image detection, audio deepfake detection, localized image manipulation detection, and text analysis. Its modular architecture enables flexible integration of heterogeneous models and supports future extensions for emerging manipulation techniques. Experimental results on representative benchmark datasets demonstrate reliable detection performance across multiple media modalities while providing a practical and user-friendly interface for forensic investigations. The platform highlights the potential of unified multimodal systems for comprehensive deepfake detection in real-world environments, and is accompanied by a public demonstrator released with this workshop submission to support reproducibility and community evaluation.

\keywords{Multimedia Forensics \and Deepfakes \and Disinformation Detection}

\end{abstract}

\section{Introduction}
\label{sec1}

Recent advances in generative artificial intelligence (AI) have significantly increased the realism and accessibility of synthetic media, including manipulated or fully generated images, videos, audio, and text. Powered by diffusion models and generative adversarial networks (GANs), such content can often be indistinguishable from authentic media, creating new challenges for trust, security, and privacy.

While these technologies offer creative and practical benefits, they also enable malicious applications such as disinformation, identity fraud, deepfake pornography, and harmful online content. The rapid spread of synthetic media on social platforms has highlighted the growing need for reliable and scalable detection systems.

In response, the media forensics community has developed a wide range of specialized detectors. However, most existing approaches focus on a single manipulation type or media modality, often rely on dataset-specific training, and frequently lack transparency or practical deployment support. This limits their applicability in real-world forensic and verification scenarios.

To address these limitations, we present the \textbf{Media Intelligence Platform (MIP)}, a unified web-based framework for multimodal media forensics. MIP integrates specialized modules for AI-generated image detection, image forgery localization, face morphing detection, deepfake video detection, audio deepfake detection, and text-based harmful content analysis. Its modular microservice architecture allows independent development and scalable deployment of heterogeneous forensic models while maintaining a consistent interface for users and external applications.

\begin{figure}[t!]
    \centering
    \includegraphics[width=1.0\linewidth]{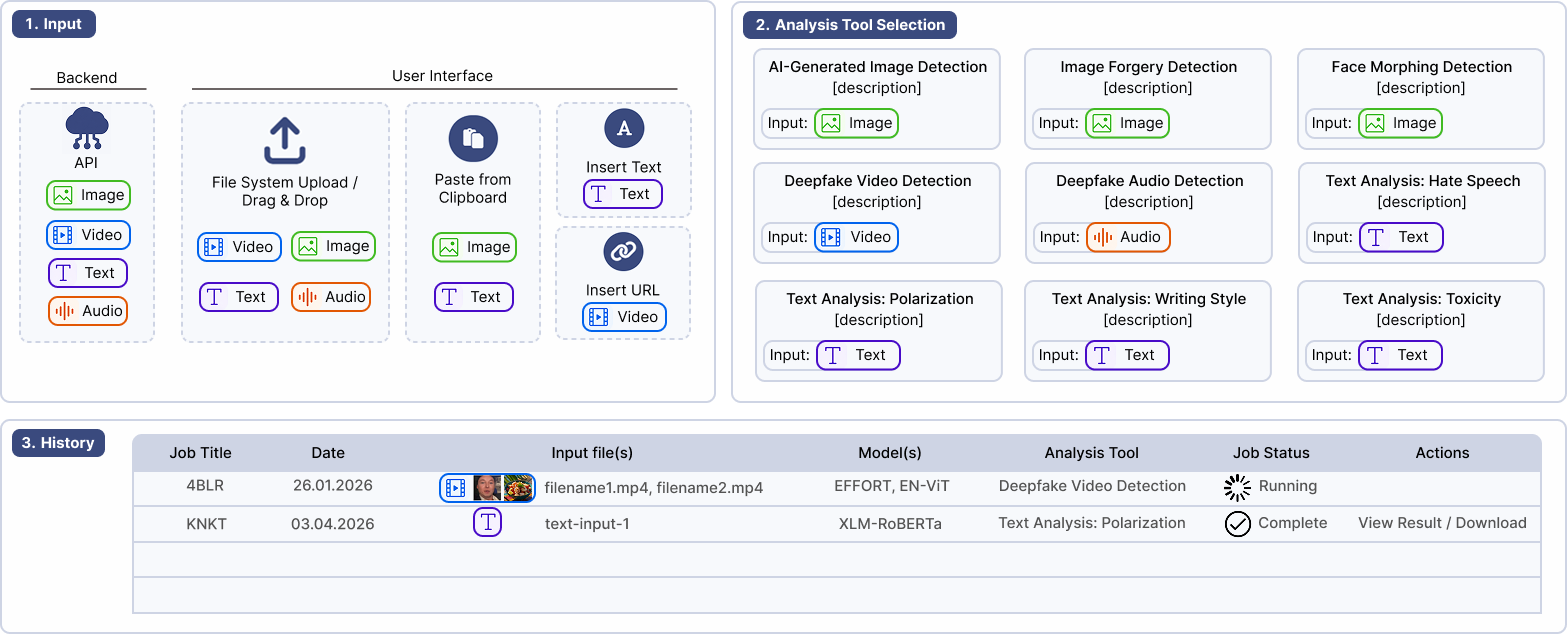}
    \caption{MIP overview. The platform supports multimodal input upload, analysis tool selection, and job history tracking.}
    \label{fig:mip-overview-input}
\end{figure}

\begin{figure}[t!]
    \centering
    \includegraphics[width=0.9\linewidth]{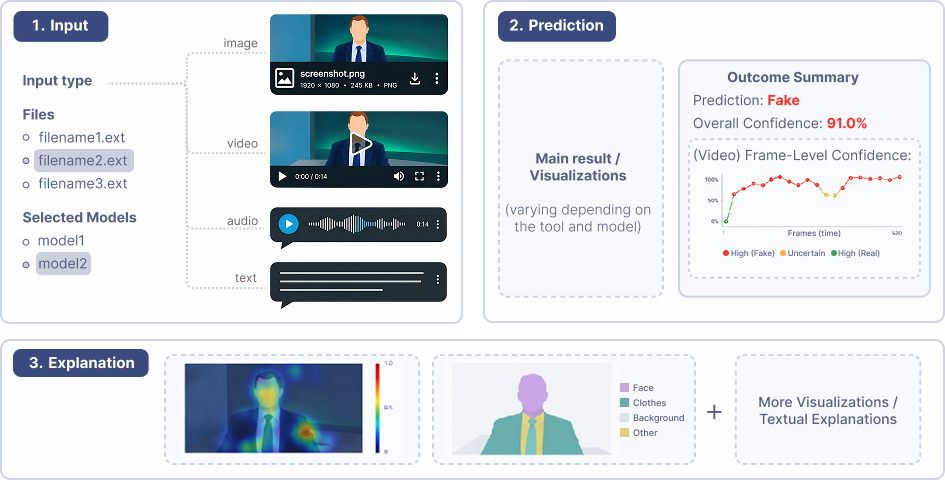}
    \caption{Example result view showing model predictions and explainable outputs.}
    \label{fig:mip-overview-results}
\end{figure}

The platform provides both a user-friendly web interface and API access for practical use in forensic investigations, media verification, and trust-and-safety applications. Experimental evaluations on representative benchmark datasets demonstrate reliable performance across all supported modalities. Fig.~\ref{fig:mip-overview-input} and Fig.~\ref{fig:mip-overview-results} illustrate the MIP interface and result visualization workflow.

\section{Related Work}\label{sec2}

\subsection{Multimodal Detection Systems}

Recent work has increasingly moved beyond single-modality detectors toward integrated verification platforms that combine multiple forensic models across images, videos, audio, and text. Browser-based systems such as InVID-WeVerify and the vera.ai Verification Plugin provide unified interfaces for media upload, URL analysis, and credibility assessment \cite{vera2026plugin,milner2026verificationassistant}. Similarly, DeepFake-O-Meter v2.0~\cite{ju2024deepfakeometer} aggregates multiple independent detectors for deepfake images, videos, and audio within a common online platform. Commercial systems such as TrueMedia~\cite{truemedia2026detector}, Sensity~\cite{sensity2026platform}, and Reality Defender~\cite{realitydefender2025sdk} extend this idea with API-based services for content moderation and fraud prevention. These systems demonstrate the growing relevance of multimodal media verification, but often suffer from limited transparency, restricted deployment options, and closed evaluation protocols.

\subsection{AI-Generated Image Detection}

Detection of AI-generated images has evolved from artifact-based methods toward more generalizable deep learning approaches. Existing methods exploit semantic inconsistencies, spatial artifacts, or generator-specific fingerprints \cite{borji2023qualitative,zhong_007101_2024,frank_007902_2020}. More recent work has shifted toward diffusion-generated images, which exhibit fewer visible artifacts and challenge traditional GAN-oriented detectors \cite{corvi_007160_2023}. Explainable AI (XAI) methods such as Grad-CAM~\cite{gradcam} and SHAP~\cite{shap} are increasingly integrated to improve transparency and support human verification.

\subsection{Image Forgery Detection and Localization}

Image forgery detection and localization focuses on identifying manipulated regions in otherwise authentic images. Traditional methods addressed specific manipulation types such as splicing~\cite{Lyu2013}, copy-move\cite{Li2013}, and inpainting~\cite{Li2017} using handcrafted forensic traces. Recent work (for an overview see~\cite{MUNAWAR2026131347}) increasingly relies on deep encoder--decoder architectures such as ManTra-Net \cite{Wu2019ManTraNet} and U-Net variants \cite{Fischinger2023DFNet}, enabling joint detection and localization of manipulated content. These approaches provide both binary authenticity decisions and pixel-level localization maps.

\subsection{Face Morphing Attack Detection}

Face morphing attack detection is typically divided into single-image (S-MAD) and differential (D-MAD) settings \cite{Antonutti2026FaceMA}. Recent methods increasingly rely on deep learning, replacing handcrafted features with discriminative embedding models. In S-MAD, self-supervised and metric-learning-based approaches have shown strong performance \cite{ivanovska_selfmad_2025,tapia_single-morphing_2025}, while D-MAD commonly employs Siamese architectures to compare probe and reference images \cite{soleymani_differential_2021}.

\subsection{Audio Deepfake Detection}

Audio deepfake detection can be divided into speech deepfake detection (DSD) and environmental sound deepfake detection (DESD). Recent methods increasingly leverage large pre-trained audio foundation models such as XLSR~\cite{sota_01_xlsr}, WavLM~\cite{Wavlm_01}, and BEATs~\cite{beats}, which are fine-tuned for downstream detection tasks. This transfer-learning paradigm has become dominant, consistently outperforming earlier handcrafted and task-specific approaches.

\subsection{Text-Based Detection}

Text-based disinformation detection increasingly combines automated classification with explainability and interactive analysis. Existing work addresses tasks such as fact-checking, propaganda detection, sentiment analysis, and rumor detection \cite{SCW19,MYP21}. More recent approaches integrate linguistic, semantic, and stylistic features to improve interpretability and robustness \cite{KAS24,LKV22}. However, most systems remain limited to isolated tasks, motivating modular architectures that combine multiple text classifiers for toxicity, hate speech, and credibility assessment. Particularly relevant to online content moderation are toxicity and hate-speech detection systems, which identify abusive, offensive, or harmful language and are increasingly deployed as building blocks within larger trust-and-safety frameworks.

\section{System Architecture and Integration Framework}\label{sec3}

The Media Intelligence Platform (MIP) is a modular, container-based framework for multimodal content analysis that exposes forensic and AI-based analysis components through a unified REST interface. The platform supports image and video forensics, audio analysis, and text classification, while remaining extensible to future modalities and detection methods. Each analysis component is independently developed and deployed as a microservice, allowing the platform to scale without modifying the surrounding infrastructure. The complete system runs on an on-premises Kubernetes cluster and provides a reference web frontend (MIP-UI) for user interaction and access management.

\begin{figure}[t]
  \centering
  \includegraphics[width=\textwidth]{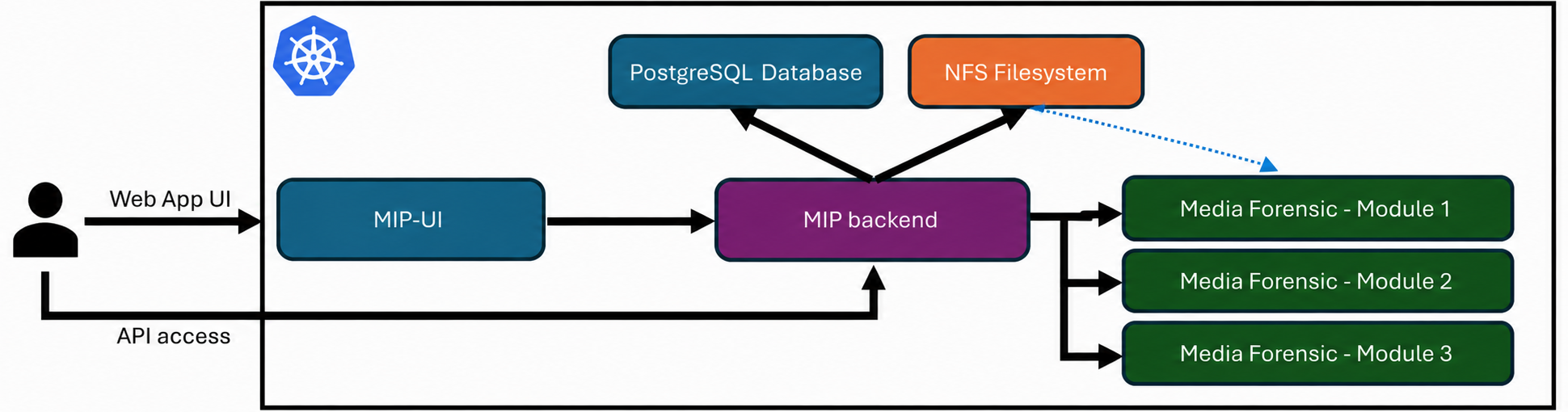}
  \caption{Schematic of the MIP microservice architecture. Modules are deployed as Kubernetes pods and communicate via standardized REST APIs. Results and metadata are stored through a shared NFS file system and PostgreSQL backend.}
  \label{fig:mip-architecture}
\end{figure}

\paragraph{Microservice Architecture}
Fig.~\ref{fig:mip-architecture} illustrates the overall architecture of MIP. We distinguish between \emph{Modules} and \emph{Models}. A \emph{Module} represents a specific analysis task (e.g.\ AI-generated image detection or text classification), while a \emph{Model} is the underlying neural network implementing the task. A single Module may encapsulate multiple Models behind a common interface, enabling interchangeable inference without exposing internal implementation details.

Each Module provides a standardized REST API for inference, status reporting, and result retrieval. Inputs can be uploaded directly or referenced via shared storage. Results are returned through a common \texttt{ModuleResult} schema, allowing uniform storage and processing across all modules. Every Module is packaged as an isolated Docker container and can optionally provide custom UI components for configuration and result visualization.

\begin{figure}[t]
  \centering
  \includegraphics[width=\textwidth]{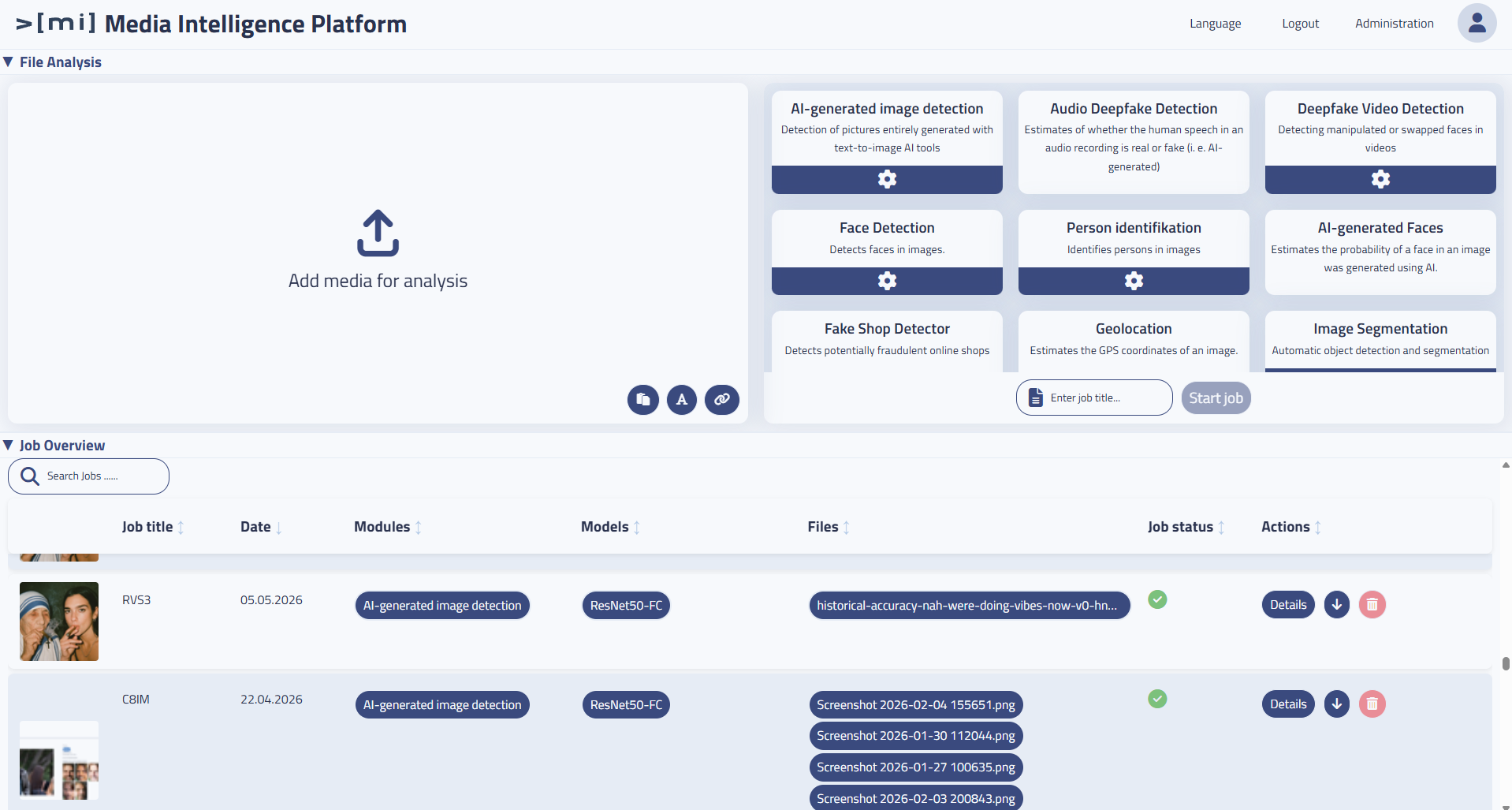}
  \caption{MIP-UI: reference frontend for module selection, job history, and result visualization.}
  \label{fig:mip-ui}
\end{figure}

\paragraph{User Interface}
The microservice API is frontend-agnostic and allows arbitrary external applications to connect to the platform. We provide MIP-UI (Fig.~\ref{fig:mip-ui}) as a reference frontend, which implements user management, role-based access control, and job tracking. The frontend dynamically lists connected Modules, renders their configuration forms, and displays module-specific result visualizations. Through its \texttt{/modules} endpoint, MIP-UI also allows external partners to build custom interfaces on top of the platform.

\paragraph{Deployment}
MIP is deployed on on-premises hardware using Kubernetes and Helm-based infrastructure-as-code. The production system runs on multi-GPU hardware, while a separate development system supports internal testing. All services operate offline, which is essential for privacy-sensitive forensic workflows.

To maximize hardware utilization, GPU resources are shared across multiple Modules using HAMi (Heterogeneous AI Computing Virtualization Middleware), enabling multiple inference services to run concurrently on a single GPU while maintaining memory isolation and fault containment.

\paragraph{Design Principles}
The platform follows four core design principles: \emph{modularity}, \emph{extensibility}, \emph{reproducibility}, and \emph{deployment independence}. These principles allow new detection modules to be integrated with minimal effort and ensure that the platform can evolve alongside emerging synthetic media threats.

\section{Visual Media Analysis Modules}\label{sec4}

\subsection{AI-Generated Image Detection}\label{sec41}

The increasing realism of text-to-image generation models has created a growing demand for reliable AI-generated image detection. In particular, passive forensic detectors are essential when proactive methods such as watermarking are absent or can be circumvented \cite{yu_006049_2021,noauthor_010000_2024}.

Within MIP, the AI-generated image detection module integrates four detector variants and 16 XAI methods. The module accepts standard image formats and allows users to compare predictions across multiple models while selecting different XAI strategies for visual interpretation of the decision process.

To improve robustness and generalization, we combine two architectural families: Convolutional Neural Networks (CNNs) and Vision Transformers (ViTs), both of which have shown strong performance for AI-image detection \cite{zhu_001000_2024,dogoulis2023improving}. Specifically, the module integrates ResNet-50 \cite{resnet} and ViT-B-16 \cite{vit} backbones with different fine-tuning strategies. While the ViT model is fully fine-tuned, the ResNet variants adapt progressively deeper layers.

Training was performed on a mixed dataset of $162$k real images from MS-COCO \cite{Lin2014} and $209$k synthetic images from the AIText2Image dataset \cite{poletti2026aigeneratedimageshumansmachines} that has been generated with Stable Diffusion XL \cite{sdxl}, DALL-E 2 \cite{dalle2} and Midjourney 5 \cite{midjourney5}. Additional augmentations such as JPEG compression and common image filters were applied to improve robustness.

Table~\ref{tab:aidetector_eval} summarizes the performance across evaluation datasets. The results show strong generalization across multiple generators unseen during training, including diffusion-based models such as FLUX, Firefly, and DALL-E 3.

\begin{table*}[h]
\centering
\caption{
F1-score (\%) comparison of AI-generated image detectors on evaluation datasets. Best results per dataset are highlighted in bold. For each dataset, the image type is reported ({\color{green}R} for real and {\color{red}F} for fake).
}
\label{tab:aidetector_eval}
\resizebox{\textwidth}{!}{%
\begin{tabular}{lcccccccc}
\toprule

\textbf{AI Detectors}
& \shortstack{\texttt{ImgN100}\cite{imagenet} \\ {\color{green}R}}
& \shortstack{\texttt{VOC12}\cite{pascal-voc-2012} \\ {\color{green}R}}
& \shortstack{\texttt{OImg-v7}\cite{openimg} \\ {\color{green}R}}
& \shortstack{\texttt{FLUX.1}\cite{flux} \\ {\color{red}F}}
& \shortstack{\texttt{Firefly}\cite{firefly} \\ {\color{red}F}}
& \shortstack{\texttt{MJ 5\&6}\cite{midjourney5, midjourney6} \\ {\color{red}F}}
& \shortstack{\texttt{DALL-E 3}\cite{dalle3} \\ {\color{red}F}}
& \shortstack{\texttt{GLIDE}\cite{glide} \\ {\color{red}F}} \\

\midrule

ResNet50[fc]
& 87.1 & 82.5 & 96.4 & 98.2 & 92.2 & \textbf{96.7} & \textbf{98.1} & 84.9 \\

ResNet50[fc+l4]
& 97.0 & 92.4 & 99.4 & \textbf{98.8} & \textbf{96.4} & 87.4 & 92.1 & 86.2 \\

ResNet50[fc+l4+l3]
& \textbf{99.2} & \textbf{96.6} & \textbf{99.6} & 95.8 & 79.0 & 89.5 & 96.2 & 74.8 \\

ViT-B-16
& 94.2 & 89.6 & 97.5 & 94.9 & 86.9 & 93.8 & 97.0 & \textbf{89.9} \\

\bottomrule
\end{tabular}%
}
\end{table*}

A distinguishing feature of this module is its support for explainability. The integrated XAI methods generate heatmaps that highlight image regions contributing most strongly to the classification decision. To further improve interpretability, these heatmaps are combined with object detection, allowing users to localize and identify objects that contain the most prominent AI-generation artifacts (see Fig.~\ref{fig:mip-overview-results}, Explanation section). This supports human analysts in understanding not only the prediction itself, but also the visual evidence behind it.

\subsection{Image Forgery Detection and Localization}\label{sec42}

The Image Forgery Detection and Localization (IFDL) module is the primary image-forensic component of MIP and complements the AI-generated image detector by targeting classical image editing operations. In contrast to deepfake detectors, which mainly focus on synthetic faces or videos, IFDL addresses manipulations such as splicing, copy-move, inpainting, object removal, and local content replacement. These manipulations remain highly relevant in real-world misinformation and fraud scenarios \cite{Guo2026rethinking}.

Traditional forensic methods rely on handcrafted traces such as JPEG artifacts, sensor noise inconsistencies, or duplicated textures. While often interpretable, their robustness decreases under strong post-processing or recompression. Recent work therefore increasingly relies on deep learning models that learn manipulation traces directly from data \cite{MUNAWAR2026131347}.

Within MIP, we integrate a hybrid U-Net–based convolutional network for joint manipulation detection and localization. The model employs a ConvNeXtV2-Base encoder pretrained on ImageNet and a U-Net–style decoder to generate high-resolution segmentation masks. In addition to pixel-level localization, a classification head predicts a global authenticity score, enabling multi-task learning of both detection and localization.

Training was performed on a mixture of authentic images from MS COCO \cite{Lin2014} and manipulated samples from DF2023 \cite{Fischinger2023DF2023} and MIML \cite{Qu_2024_CVPR}, with standard augmentation to improve robustness. The model jointly learns to (i) classify images as authentic or manipulated and (ii) localize suspicious regions.

\begin{figure}[t]
    \centering
    \includegraphics[width=1.0\linewidth]{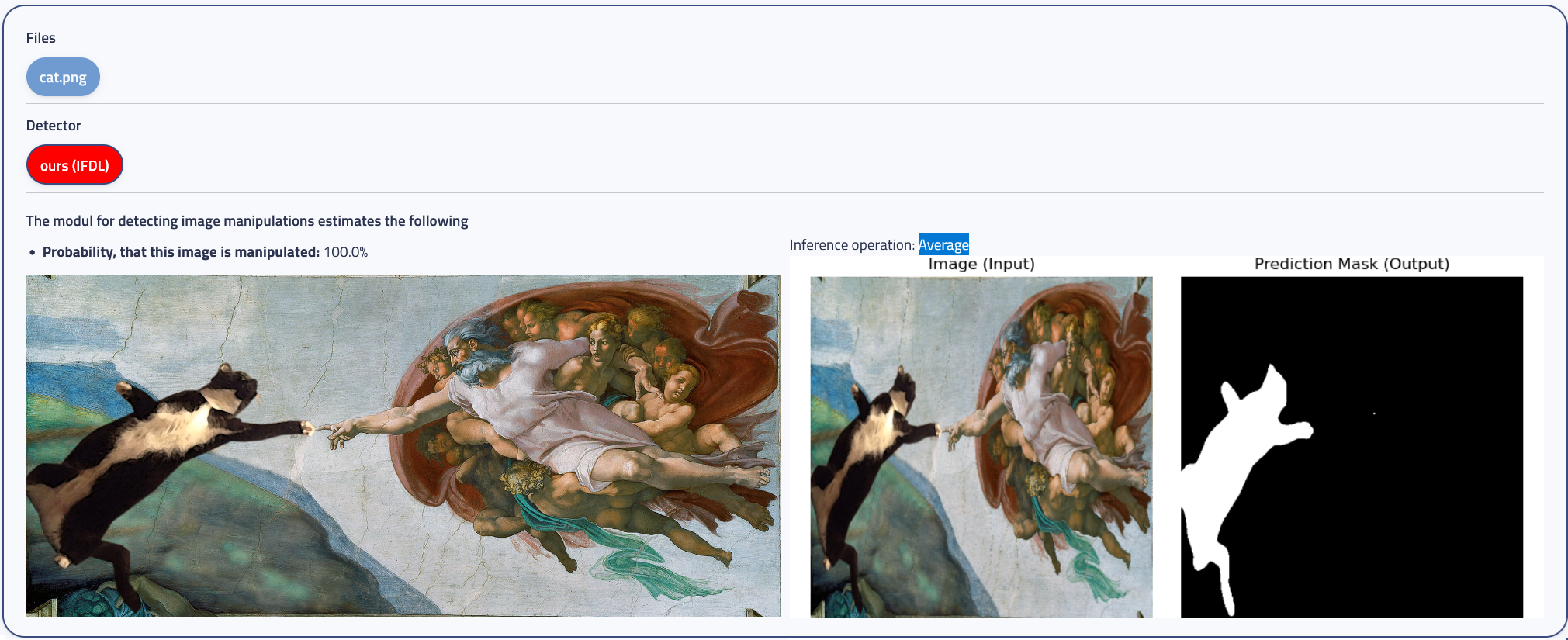}
    \caption{Example output of the IFDL module. The manipulated input image (left) is assigned a high manipulation probability, while the predicted localization mask (right) highlights the tampered region.}
    \label{fig:ifdlcat}
\end{figure}

Benchmark evaluation in Table~\ref{tab:ifdlevaluation} shows that the proposed model consistently outperforms previous state-of-the-art approaches across multiple forensic datasets, particularly when using test-time augmentation. This demonstrates strong generalization across diverse manipulation types and acquisition conditions.

\begin{table*}[t]
\centering
\caption{Comparison of state-of-the-art approaches and our proposed IFDL-Model in default mode respectively averaging over the results of the augmented (horizontally/vertically mirrored, rotated) input image. Highest values are highlighted in bold.}
\resizebox{\textwidth}{!}{%
\begin{tabular}{lcccccccccccccccc}
\toprule

& \multicolumn{16}{c}{Test Datasets} \\
\cmidrule(lr){2-17}

Models
& \multicolumn{3}{c}{\texttt{CASIA} \cite{CASIA}}
& \multicolumn{3}{c}{\texttt{Columbia}\cite{Columbia}}
& \multicolumn{3}{c}{\texttt{DSO}\cite{DSO}}
& \multicolumn{3}{c}{\texttt{NIST}\cite{NIST}}
& \multicolumn{4}{c}{\textbf{Average}} \\
\cmidrule(lr){2-4}
\cmidrule(lr){5-7}
\cmidrule(lr){8-10}
\cmidrule(lr){11-13}
\cmidrule(lr){14-17}

& AUC & F1 & IoU
& AUC & F1 & IoU
& AUC & F1 & IoU
& AUC & F1 & IoU
& AUC & F1 & IoU & \textbf{Mean} \\
\midrule

DFCN~\cite{Zhuang2021}
& 0.654 & 0.192 & 0.119
& 0.789 & 0.541 & 0.395
& 0.724 & 0.303 & 0.227
& 0.778 & 0.250 & 0.204
& 0.736 & 0.322 & 0.236 & 0.431 \\

FSim~\cite{Mayer2019ForSim}
& 0.554 & 0.169 & 0.102
& 0.731 & 0.604 & 0.474
& 0.796 & 0.487 & 0.371
& 0.642 & 0.188 & 0.123
& 0.681 & 0.362 & 0.268 & 0.437 \\

MNet~\cite{Wu2019ManTraNet}
& 0.776 & 0.130 & 0.086
& 0.747 & 0.357 & 0.258
& 0.795 & 0.344 & 0.253
& 0.634 & 0.088 & 0.054
& 0.738 & 0.230 & 0.163 & 0.377 \\

Wu22~\cite{Wu2022OSN}
& 0.873 & 0.509 & 0.465
& 0.862 & 0.707 & 0.608
& 0.854 & 0.436 & 0.308
& 0.783 & 0.332 & 0.255
& 0.843 & 0.496 & 0.409 & 0.583 \\

DF-Net~\cite{Fischinger2023DFNet}
& 0.906 & 0.589 & 0.496
& 0.880 & 0.757 & 0.679
& 0.769 & 0.360 & 0.246
& \textbf{0.793} & 0.299 & 0.226
& 0.837 & 0.501 & 0.411 & 0.583 \\


\textbf{ours}
& 0.907 & 0.705 & 0.639
& 0.983 & 0.960 & 0.945
& 0.831 & 0.611 & 0.508
& 0.761 & 0.423 & 0.346
& 0.871 & 0.675 & 0.609 & 0.718 \\

\textbf{ours (AVG) }
& \textbf{0.917} & \textbf{0.725} & \textbf{0.661}
& \textbf{0.991} & \textbf{0.962} & \textbf{0.951}
& \textbf{0.864} & \textbf{0.678} & \textbf{0.576}
& 0.766 & \textbf{0.428} & \textbf{0.352}
& \textbf{0.884} & \textbf{0.698} & \textbf{0.635} & \textbf{0.739} \\

\bottomrule
\end{tabular}%
}
\label{tab:ifdlevaluation}
\end{table*}

Within MIP, the IFDL module returns both a manipulation confidence score and a localization mask, enabling transparent and human-interpretable forensic analysis (see Fig.~\ref{fig:ifdlcat} for an example). This complements the AI-generated image detector by capturing manipulations applied to otherwise authentic photographs. Combining both signals broadens the range of detectable manipulations and improves robustness in practical forensic workflows.

Future work will focus on improving robustness against unseen editing tools, strong compression, and adversarial post-processing, as well as integrating forensic evidence with metadata and cross-modal reasoning.

\subsection{Face Morphing Detection}\label{sec45}
Face morphing combines the facial characteristics of two or more individuals into a single synthetic portrait. While originally developed for entertainment purposes, face morphing has become a serious security concern because morphed passport images can deceive face recognition systems and allow multiple individuals to authenticate against the same identity document \cite{ferrara_magic_2014}. This attack scenario, known as a \emph{face morphing attack}, represents an important threat for border control, identity verification, and biometric security systems \cite{ambikapathy_biometric_2024}.

Recent morphing attack detection methods increasingly rely on deep learning, replacing handcrafted features with discriminative feature extractors pretrained on face recognition tasks. Within MIP, the deployed morphing attack detector is based on ArcFace~\cite{deng_arcface_2019} with a ResNet-100 backbone. ArcFace learns highly discriminative facial embeddings by enforcing angular separation between identities, which makes it particularly suitable for detecting morphed images that contain blended identity information.

For training, we curated a diverse morphing dataset from multiple passport-like face datasets and generated morphs using both classical landmark-based and recent deep learning–based morphing methods. This setup provides broad variability in acquisition conditions and morph generation techniques, improving generalization to unseen attacks.

The model was fine-tuned for binary classification (bona fide vs.\ morphed) while preserving the discriminative facial representations learned during large-scale face recognition pretraining.

\begin{table*}[ht!]
\centering
\caption{Equal Error Rate (EER \%) obtained by the evaluated methods on the different morph generation techniques. From left to right the generation methods are: DiffMorpher, Greedy-DiM, MipGAN-II, OpenCV, StyleGAN2 and UTW.}
\begin{tabular}{|l|c|c|c|c|c|c||c|}
\hline
Algorithm & DM & G-DiM & MGAN-II & OCV & SGAN2 & UTW & \textbf{Avg.} \\
\hline
SelfMAD \cite{ivanovska_selfmad_2025} & 51.45 & 5.45 & 2.90 & \textbf{0.59} & 73.25 & 52.98 & 31.10 \\
SelfMAD++ \cite{IVANOVSKA2026103921} & 48.22 & 25.04 & 2.04 & 0.85 & 46.68 & 14.48 & 22.89 \\
SPL-MAD \cite{fang_unsupervised_2022} & 44.32 & \textbf{4.38} & 20.59 & 17.75 & 55.62 & 80.47 & 37.19 \\
CrEffViT \cite{effvit} & 18.43 & 42.45 & 18.29 & 22.98 & 12.29 & 17.24 & 21.95 \\
FatFormer \cite{liu_forgery-aware_2024} & 16.31 & 10.87 & 1.19 & 6.80 & 12.43 & 12.50 & 10.02 \\
Triplet-Loss \cite{tapia_single-morphing_2025} & 11.85 & 5.33 & \textbf{0.98} & 5.10 & 11.56 & \textbf{0.70} & 5.92 \\
\textbf{ArcFace \cite{deng_arcface_2019}} & \textbf{3.79} & 4.66 & 3.05 & 4.24 & \textbf{3.77} & 1.99 & \textbf{3.58} \\
\hline
\end{tabular}
\label{tab:eer_standard}
\end{table*}

Table~\ref{tab:eer_standard} summarizes the evaluation results. The ArcFace-based detector achieves the lowest average Equal Error Rate (EER) of \textbf{3.58\%} across all evaluated morph generation methods, outperforming recent morphing attack detectors and demonstrating strong robustness against both classical and learning-based morphs.

Within MIP, the Face Morphing Detection module accepts portrait images and returns a morphing probability through the standardized interface. This allows seamless integration with the remaining visual forensic modules, enabling comprehensive authenticity assessment of facial images and identity documents.

Future work will focus on improving robustness against emerging diffusion-based morphing techniques, which increasingly reduce visible blending artifacts and pose new challenges for forensic analysis.

\FloatBarrier
\subsection{Face Swap Detection in Videos}\label{sec46}

Face swap is a manipulation technique that transfers the face of a source person onto a target person in images or videos. When performed frame-wise using deep learning models, the resulting manipulated videos are commonly referred to as \emph{deepfakes}. The most common manipulation categories include face swap, face reenactment, lip-sync, and full face synthesis.

The deepfake video module in MIP focuses on detecting face swap and face reenactment manipulations. Given an input video, the system analyzes each detected face across sampled frames and assigns a manipulation score. The output is a per-person timeline of confidence values, enabling both frame-level and video-level forensic inspection.

Rather than proposing a new detector, we integrate \textbf{19 state-of-the-art deepfake detection methods} into a unified benchmark framework following DeepFakeBench \cite{yan2023deepfakebench}. This allows direct comparison under identical preprocessing and inference conditions. For evaluation, we curated a benchmark of 1000 videos (50 real and 50 fake from each of 10 datasets), enabling cross-dataset performance analysis.

\renewcommand{\arraystretch}{0.2}
\begin{table*}[t]
\centering
\caption{Deepfake detection methods available on the Media-Intelligence-Platform. Performance on benchmark dataset, sorted by descending video AUC.}
\label{tab:detectors}
\resizebox{\textwidth}{!}{%
\begin{tabular}{llrc}
\toprule
\textbf{Detector Method} & \textbf{Category} & \textbf{Video AUC} & \textbf{Reference} \\
\midrule

\textbf{EFFORT} & \textbf{ViT, CLIP} & \textbf{0.928} & \cite{yan_orthogonal_2025} \\

\textbf{selimsef FbF} & \textbf{CNN, Ensemble} & \textbf{0.917} & \cite{seferbekov_selimsefdfdc_deepfake_challenge_2026} \\

\textbf{GenConvViT} & \textbf{CNN, ViT} & \textbf{0.908} & \cite{wodajo2023deepfake} \\

CC ViT & CNN, ViT & 0.903 & \cite{effvit} \\

Efficient ViT & CNN, ViT & 0.869 & \cite{effvit} \\

UCF & CNN, Disentanglement, Autoencoder & 0.864 & \cite{yan_ucf_2023} \\

Xception & CNN & 0.858 & \cite{rossler_faceforensics_2019} \\

ID-unaware & CNN, Artifact Detection & 0.845 & \cite{dong_02_2023} \\

CORE & CNN, Representation Learning & 0.843 & \cite{ni_core_2022} \\

RECCE & CNN, Autoencoder, Graph & 0.840 & \cite{cao_end--end_2022} \\

F3Net & CNN, Frequency & 0.839 & \cite{qian_thinking_2020} \\

FFD & CNN, Attention & 0.839 & \cite{dang_detection_2020} \\

SPSL & CNN, Frequency & 0.838 & \cite{liu_spatial-phase_2021} \\

SRM & CNN, Frequency, Attention & 0.832 & \cite{luo_generalizing_2021} \\

CapsuleNet & CNN, Capsule Network & 0.795 & \cite{nguyen_capsule-forensics_2019} \\

EfficientNet-B4 & CNN & 0.765 & \cite{tan_efficientnet_2020} \\

ResNet34 & CNN & 0.725 & \cite{resnet} \\

MesoNet4Inception & CNN, Inception & 0.720 & \cite{afchar_mesonet_2018} \\

LRNet & RNN, Landmark, Temporal & 0.639 & \cite{sun_improving_2021} \\

\bottomrule
\end{tabular}%
}
\end{table*}
\renewcommand{\arraystretch}{1.0}

The benchmark results (Table~\ref{tab:detectors}) show that the strongest detectors are based on large pretrained vision backbones such as CLIP and hybrid CNN--Transformer architectures. The best-performing method, EFFORT, achieves a video AUC of \textbf{0.928}, followed by Selimsef FbF (0.917) and GenConvViT (0.908).

Our analysis revealed three important findings.

\paragraph{Preprocessing quality.}
All detectors depend on accurate face detection and tracking. Weak preprocessing pipelines frequently cause missed faces, incorrect identity assignment, or unstable tracking, significantly degrading detection performance. By replacing the preprocessing stage with a unified GPU-accelerated pipeline, we improved runtime by up to $100\times$ for some detectors while increasing robustness across all integrated methods.

\paragraph{Identity leakage.}
Cross-dataset generalization remains a major challenge. As shown in \cite{dong_02_2023}, many detectors implicitly learn facial identities instead of manipulation artifacts, causing strong in-dataset performance but poor transferability. Identity-disjoint train/test splits are therefore essential for reliable evaluation.

\paragraph{Foundation model adaptation.}
The best-performing methods consistently rely on large pretrained vision models and lightweight fine-tuning strategies such as LoRA, SVD-based adaptation, or partial layer unfreezing. These approaches preserve general visual representations while adapting to forensic artifacts.

\begin{figure}[h!]
  \centering
  \includegraphics[width=1.0\textwidth]{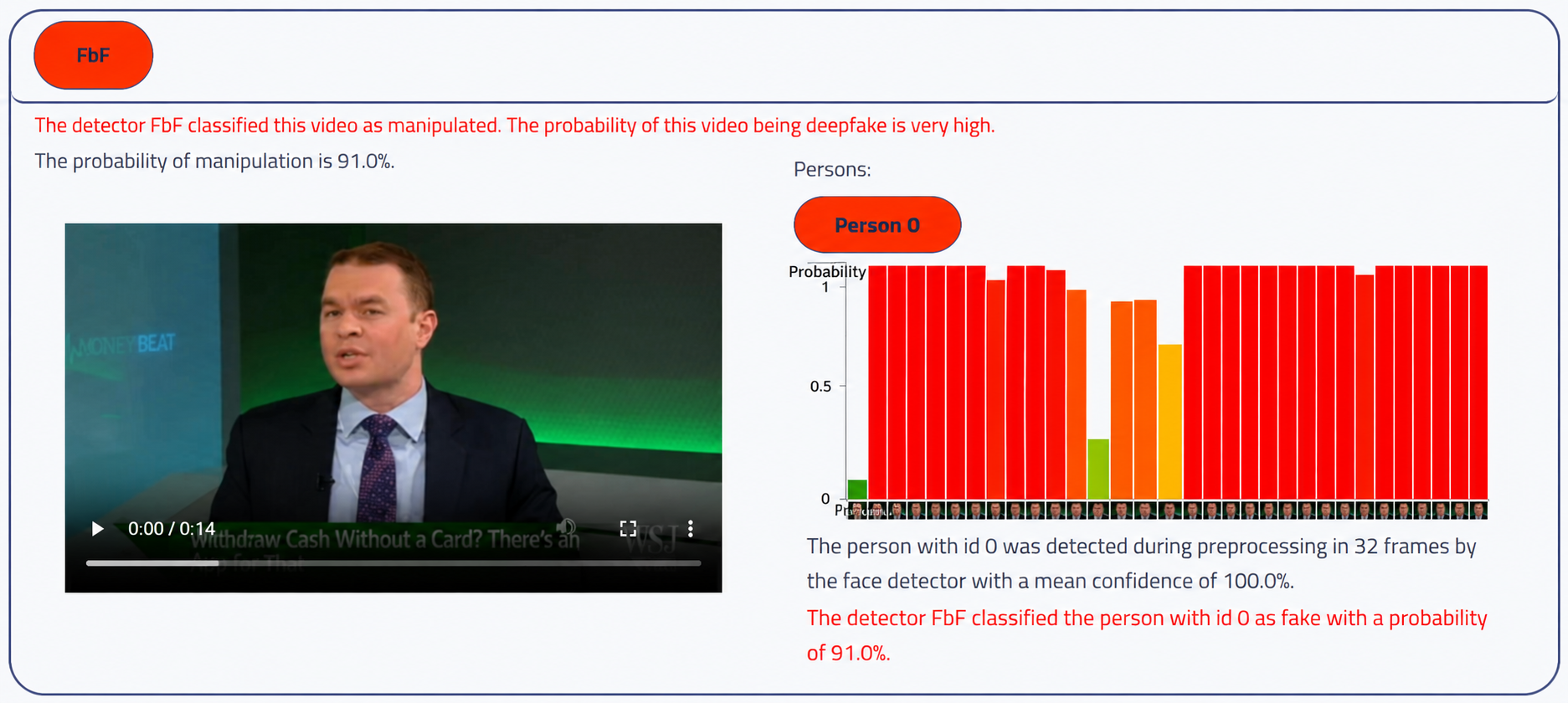}
  \caption{For each video and each detector result the report will display a bar chart for each person with the predicted faceswap detection confidence per frame or sequence.}
  \label{fig:mip-faceswap}
\end{figure}

\FloatBarrier

Within MIP, users can select any of the 19 integrated detectors for video analysis. For each tracked individual, the system visualizes frame-wise manipulation scores and aggregated video-level predictions (Fig.~\ref{fig:mip-faceswap}), enabling direct comparison of detector behavior and supporting interpretable forensic analysis.

Future work will extend this framework toward emerging multimodal diffusion-based video generation models, where the forensic task remains the same: determining whether facial content originates from authentic recordings or synthetic generation.

\FloatBarrier
\section{Audio Deepfake Detection}\label{sec5}

Audio deepfake detection has become increasingly important with the rapid progress of neural speech synthesis and voice cloning models. Synthetic speech can be used for impersonation, fraud, and disinformation, making robust forensic detection essential.

The Media Intelligence Platform provides a dedicated Deepfake Speech Detection (DSD) module for analyzing manipulated speech recordings. Following recent trends in the literature, the module is based on a fine-tuned pre-trained audio foundation model.

\begin{figure}[h]
    \centering
    \includegraphics[width=1.0\textwidth]{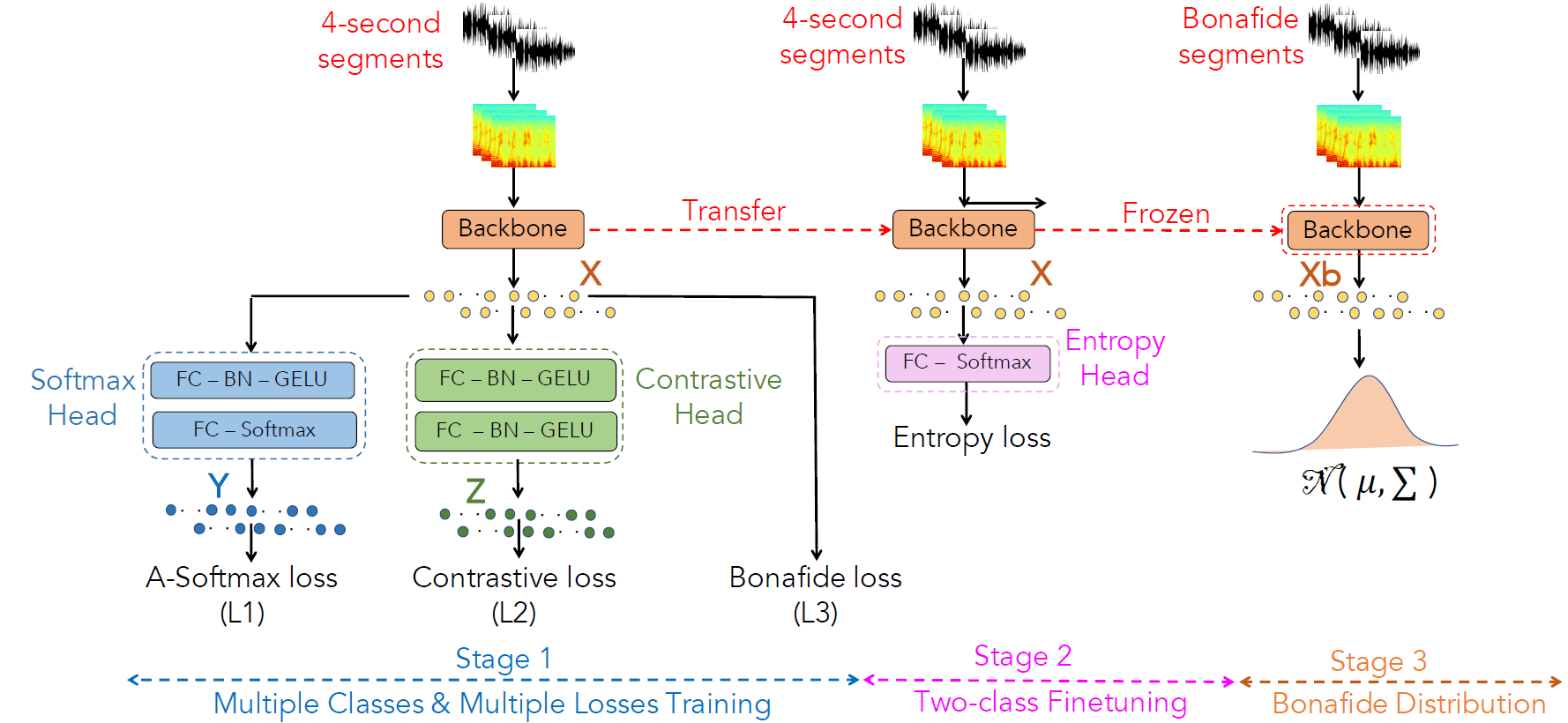}
    \caption{Overview of the Deepfake Speech Detection (DSD) pipeline. The model uses a three-stage training strategy based on feature extraction, discriminative embedding learning, and Mahalanobis-based inference.}
    \label{fig:lam_f02}
\end{figure}

\paragraph{Model Architecture.}
The DSD module uses the pretrained WavLM backbone~\cite{Wavlm_01} to extract feature embeddings from 4-second audio segments. Training follows a three-stage optimization strategy (see Fig.~\ref{fig:lam_f02}). First, multiple supervisory objectives are applied jointly, including A-Softmax loss for multi-class spoof classification, Contrastive Loss to separate bona fide and spoofed embeddings, and Center Loss to compact bona fide embeddings. In the second stage, the model is fine-tuned for binary classification. In the final stage, the backbone is frozen and bona fide embeddings are used to estimate a Mahalanobis distribution, enabling anomaly-based inference for unseen spoofing attacks.

\paragraph{Training and Evaluation.}
The model was trained on ASVspoof 2019 (LA)~\cite{ASV19}, ASVspoof 2021 (LA \& DF)~\cite{ASV21}, and ASVspoof5~\cite{ASV24}. Evaluation was performed on three unseen datasets: In-The-Wild~\cite{intwi_ds}, ASVspoof5 Evaluation Subset~\cite{ASV24}, and FakeAVCeleb~\cite{khalid2021fakeavceleb}, allowing assessment of cross-dataset generalization.

\begin{table}[h]
\centering
\caption{Cross-dataset performance of the DSD module.}
\label{table:lam_res02}
\renewcommand{\arraystretch}{0.85} 
\begin{tabular}{lccc}
\toprule
\textbf{Test Dataset} & \textbf{Accuracy} & \textbf{F1} & \textbf{AUC} \\
\midrule
In-The-Wild & 0.87 & 0.79 & 0.83 \\
ASVspoof5 Eval & 0.91 & 0.91 & 0.92 \\
FakeAVCeleb & 0.99 & 0.99 & 0.99 \\
\bottomrule
\end{tabular}
\end{table}

The results demonstrate strong generalization across unseen datasets, indicating robustness against a wide range of synthetic speech generation methods.

\paragraph{MIP Integration.}
Within MIP, the DSD module accepts uploaded speech recordings and returns segment-wise and recording-level authenticity predictions. This enables forensic inspection of manipulated speech and complements the platform’s visual and textual analysis modules for multimodal media verification.

\section{Text Analysis Modules}\label{sec6}

Text-based disinformation detection cannot usually be reduced to a simple binary true/false classification problem. Instead, misleading content often manifests through multiple linguistic, semantic, stylistic, and ideological signals. To address this, MIP integrates a modular text analysis framework that combines several specialized classifiers for detecting harmful or manipulative communication patterns.

\paragraph{Modular Classification Framework.}
The framework is built around transformer-based language models, primarily XLM-RoBERTa (XLM-R)~\cite{XLM-R}, fine-tuned for a range of classification tasks on German news and social media datasets. The implemented modules cover hate speech, toxicity, sexism, extremism, offensive language, sentiment, reporting style, and writing style detection.

To improve interpretability and robustness, transformer embeddings can be combined with handcrafted linguistic features such as punctuation patterns, sentiment scores, stylistic indicators, and structural text properties. These features complement contextual embeddings by capturing rhetorical and surface-level signals relevant to manipulative communication.

Experimental results show that transformer-based models perform strongly on standard classification tasks, while fusion architectures improve robustness in multilabel and long-document settings.

\paragraph{Disinformation Signals.}
Rather than directly classifying content as ``fake'' or ``real'', the framework focuses on identifying indicators commonly associated with disinformation and manipulation. These include propaganda framing, clickbait, sensationalism, polarization, subjective writing, toxicity, and ideological extremism. Such signals often become more informative when analyzed jointly rather than in isolation.

\paragraph{Rule-Based Aggregation.}
Classifier outputs are aggregated into four reliability levels (\emph{Highly Questionable}, \emph{Questionable}, \emph{Slightly Questionable}, \emph{Reliable}) using weighted rules based on indicator severity and co-occurrence, providing an interpretable overall assessment.
This aggregation combines detector outputs using weighted rules based on the severity and co-occurrence of individual indicators, allowing a more transparent and explainable reliability assessment.

\paragraph{MIP Integration.}
Within MIP, users can run individual text classifiers independently or combine them into a unified disinformation analysis pipeline. This modular setup supports applications such as content moderation, propaganda analysis, journalism support, and credibility assessment, extending the platform beyond multimedia forensics toward broader multimodal media intelligence.

\section{Publicly Available Demonstrator }\label{sec7}

To support reproducibility and enable exploration of the Media Intelligence Platform (MIP), we provide a publicly accessible demonstration instance at \url{https://media-intelligence.ait.ac.at/}.

The demonstrator provides access to multimodal analysis modules for images, videos, audio, and text, including AI-generated image detection, image manipulation localization, face morphing detection, deepfake video and audio detection, hate speech detection, and text toxicity detection.
Readers can evaluate these modules using the demo account (\texttt{mip-demo} / \texttt{spot\_the\_fakes2026}). 

\FloatBarrier
\section{Conclusion}\label{sec8}

In this paper, we presented the Media Intelligence Platform (MIP), a unified and modular framework for multimodal media forensics. Unlike most existing approaches that focus on a single manipulation type or modality, MIP integrates state-of-the-art analysis modules for AI-generated images, image forgery localization, face morphing detection, deepfake video and audio detection, and text-based content analysis within a common microservice-based architecture.

The proposed design enables scalable deployment, standardized access through REST APIs, and flexible integration of heterogeneous forensic models. By combining multiple complementary analysis modules, MIP reduces tool fragmentation and supports practical forensic workflows involving images, videos, audio, and text.
Experimental evaluations across representative benchmark datasets demonstrate that the integrated modules achieve strong performance across their respective tasks. The public demonstrator further provides researchers and practitioners with direct access to the platform for experimentation and validation.

Future work will focus on extending the platform to emerging manipulation techniques, improving robustness against unseen generative models, and strengthening multimodal reasoning across heterogeneous forensic evidence.

\section*{Acknowledgements}
This work has received funding from the European Union’s Horizon Europe research
and innovation programme under GA Nº 10112169 (SafeTravellers). This work was also supported by the GADMO 2.0 project, co-funded by the European Union’s Digital Europe Programme (DIGITAL) and the Austrian Research Promotion Agency (FFG), under grant agreement No. 101083573.

%
%
\bibliographystyle{splncs04}
\bibliography{main}

@String(CVPR  = {IEEE Conf. Comput. Vis. Pattern Recog.})

@String(ICCV  = {Int. Conf. Comput. Vis.})

@String(ECCV  = {Eur. Conf. Comput. Vis.})

@String(CVPRW = {IEEE Conf. Comput. Vis. Pattern Recog. Worksh.})

@String(AAAI  = {AAAI})

@String(ICPR  = {Int. Conf. Pattern Recog.})

@String(ICASSP=	{ICASSP})

@String(CVPR  = {CVPR})

@String(ICCV  = {ICCV})

@String(ECCV  = {ECCV})

@String(CVPRW = {CVPRW})

@String(ICPR  = {ICPR})

@InProceedings{beats,
  title = 	 {{{BEAT}s: Audio Pre-Training with Acoustic Tokenizers}},
  author =       {Chen, Sanyuan and Wu, Yu and Wang, Chengyi and Liu, Shujie and Tompkins, Daniel and Chen, Zhuo and Che, Wanxiang and Yu, Xiangzhan and Wei, Furu},
  booktitle = 	 {Proceedings of the 40th International Conference on Machine Learning},
  pages = 	 {5178--5193},
  year = 	 {2023},
  editor = 	 {Krause, Andreas and Brunskill, Emma and Cho, Kyunghyun and Engelhardt, Barbara and Sabato, Sivan and Scarlett, Jonathan},
  volume = 	 {202},
  series = 	 {Proceedings of Machine Learning Research},
  month = 	 {23--29 Jul},
  publisher =    {PMLR},
  address = {Honolulu, Hawaii, USA},
  url = 	 {https://proceedings.mlr.press/v202/chen23ag.html}
}

@ARTICLE{Wavlm_01,
  author={Chen, Sanyuan and Wang, Chengyi and Chen, Zhengyang and Wu, Yu and Liu, Shujie and Chen, Zhuo and Li, Jinyu and Kanda, Naoyuki and Yoshioka, Takuya and Xiao, Xiong and Wu, Jian and Zhou, Long and Ren, Shuo and Qian, Yanmin and Qian, Yao and Wu, Jian and Zeng, Michael and Yu, Xiangzhan and Wei, Furu},
  journal={IEEE Journal of Selected Topics in Signal Processing}, 
  title={{WavLM: Large-Scale Self-Supervised Pre-Training for Full Stack Speech Processing}},
  year={2022},
  volume={16},
  number={6},
  pages={1505-1518},
  doi={10.1109/JSTSP.2022.3188113}}

@misc{ASV24,
  title={{The ASVspoof 2024 Challenge}},
  year={2024},
  howpublished={\url{https://www.asvspoof.org/}},
}

@inproceedings{khalid2021fakeavceleb,
  title={{FakeAVCeleb: A Novel Audio-Video Multimodal Deepfake Dataset}},
  author={Khalid, Hasam and Tariq, Shahroz and Kim, Minha and Woo, Simon S},
  booktitle={Thirty-fifth Conference on Neural Information Processing Systems Datasets and Benchmarks Track (Round 2)},
  year={2021}
}

@inproceedings{intwi_ds,
author = {Müller, Nicolas and Czempin, Pavel and Dieckmann, Franziska and Froghyar, Adam and Böttinger, Konstantin},
year = {2022},
month = {03},
pages = {},
title = {{Does Audio Deepfake Detection Generalize?}},
doi = {10.48550/arXiv.2203.16263}
}

@inproceedings{ASV21,
  title={{ASVspoof 2021: accelerating progress in spoofed and deepfake speech detection}},
  author={Junichi Yamagishi and Xin Wang and Massimiliano Todisco and Md Sahidullah and Jose Patino and Andreas Nautsch and Xuechen Liu and Kong Aik Lee and Tomi Kinnunen and Nicholas Evans and Héctor Delgado},
  booktitle={Workshop-Automatic Speaker Verification and Spoofing Coutermeasures Challenge (ASVspoof)},
  year={2021}
}

@article{ASV19,
author = {Wang, Xin and Yamagishi, Junichi and Todisco, Massimiliano and Delgado, Héctor and Nautsch, Andreas and Evans, Nicholas and Sahidullah, Md and Vestman, Ville and Kinnunen, Tomi and Lee, Kong Aik and Juvela, Lauri and Alku, Paavo and Peng, Yu-Huai and Hwang, Hsin-Te and Tsao, Yu and Wang, Hsin-min and Le Maguer, Sébastien and Becker, Markus and Henderson, Fergus and Ling, Zhen-Hua},
year = {2020},
month = {05},
pages = {101114},
title = {{ASVspoof 2019: A large-scale public database of synthetized, converted and replayed speech}},
volume = {64},
journal = {Computer Speech \& Language},
doi = {10.1016/j.csl.2020.101114}
}

@article{MUNAWAR2026131347,
title = {{Deployable Image Forgery Detection and Localization: A Comprehensive Review and Practitioner´s Guide}},
journal = {Expert Systems with Applications},
volume = {311},
pages = {131347},
year = {2026},
issn = {0957-4174},
doi = {https://doi.org/10.1016/j.eswa.2026.131347},
url = {https://www.sciencedirect.com/science/article/pii/S0957417426002605},
author = {Maryam Munawar and Mourad Oussalah},
}

@inproceedings{Wu2019ManTraNet,
author={Wu, Yue and AbdAlmageed, Wael and Natarajan, Premkumar},
booktitle={2019 IEEE/CVF Conference on Computer Vision and Pattern Recognition (CVPR)},
title={{ManTra-Net: Manipulation Tracing Network for Detection and Localization of Image Forgeries With Anomalous Features}},
year={2019},
pages={9535-9544},
doi={10.1109/CVPR.2019.00977}}

@ARTICLE{Zhuang2021,
  author={Zhuang, Peiyu and Li, Haodong and Tan, Shunquan and Li, Bin and Huang, Jiwu},
  journal={IEEE Transactions on Information Forensics and Security}, 
  title = {{Image Tampering Localization Using a Dense Fully Convolutional Network}},
  year={2021},
  volume={16},
  number={},
  pages={2986-2999},
  doi={10.1109/TIFS.2021.3070444}}

@article{Wu2022OSN,
	title = {{Robust Image Forgery Detection against Transmission over Online Social Networks}},
	issn = {1556-6021},
	doi = {10.1109/TIFS.2022.3144878},
    author = {Wu, Haiwei and Zhou, Jiantao and Tian, Jinyu and Liu, Jun and Qiao, Yu},
	journal = {{IEEE} Transactions on Information Forensics and Security},
	year = {2022}
	}

@article{Li2013,
author = {Li, L. and Li, S. and Zhu, Hancheng and Chu, Shu-Chuan and Roddick, John and Pan, Jeng-Shyang},
year = {2013},
month = {01},
pages = {46-56},
title = {{An Efficient Scheme for Detecting Copy-Move Forged Images by Local Binary Patterns}},
volume = {4},
journal = {Journal of Information Hiding and Multimedia Signal Processing}
}

@article{Lyu2013,
author = {Lyu, Siwei and Pan, Xunyu and Zhang, Xing},
year = {2013},
month = {11},
pages = {202-221},
title = {{Exposing Region Splicing Forgeries with Blind Local Noise Estimation}},
volume = {110},
journal = {International Journal of Computer Vision},
doi = {10.1007/s11263-013-0688-y}
}

@ARTICLE{Li2017,
  author={Li, Haodong and Luo, Weiqi and Huang, Jiwu},
  journal={IEEE Transactions on Information Forensics and Security}, 
  title={{Localization of Diffusion-Based Inpainting in Digital Images}}, 
  year={2017},
  volume={12},
  number={12},
  pages={3050-3064},
  doi={10.1109/TIFS.2017.2730822}}

@InProceedings{Lin2014,
author="Lin, Tsung-Yi
and Maire, Michael
and Belongie, Serge
and Hays, James
and Perona, Pietro
and Ramanan, Deva
and Doll{\'a}r, Piotr
and Zitnick, C. Lawrence",
editor="Fleet, David
and Pajdla, Tomas
and Schiele, Bernt
and Tuytelaars, Tinne",
title={{Microsoft COCO: Common Objects in Context}},
booktitle="Computer Vision -- ECCV 2014",
year="2014",
publisher="Springer International Publishing",
address="Cham",
pages="740--755",
isbn="978-3-319-10602-1"
}

@INPROCEEDINGS{Qu_2024_CVPR,
  author={Qu, Chenfan and Zhong, Yiwu and Liu, Chongyu and Xu, Guitao and Peng, Dezhi and Guo, Fengjun and Jin, Lianwen},
  booktitle={2024 IEEE/CVF Conference on Computer Vision and Pattern Recognition (CVPR)}, 
  title     = {{Towards Modern Image Manipulation Localization: A Large-Scale Dataset and Novel Methods}},
  year={2024},
  volume={},
  number={},
  pages={10781-10790},
  doi={10.1109/CVPR52733.2024.01025}}

@article{Fischinger2023DF2023,
	title = {{DF2023: The Digital Forensics 2023 Dataset for Image Forgery Detection}},
    author = {Fischinger, David and Boyer, Martin},
	journal = {Irish Machine Vision and Image Processing conference},
	year = 2023
}

@article{Fischinger2023DFNet,
author = {Fischinger, David and Boyer, Martin},
year = {2023},
title = {{DF-Net: The Digital Forensics Network for Image Forgery Detection.}},
journal = {Irish Machine Vision and Image Processing conference},
}

@article{Guo2026rethinking,
  title        = {{Rethinking VLMs for Image Forgery Detection and Localization}},
  author       = {Guo, Shaofeng and Cui, Jiequan and Hong, Richang},
  journal      = {arXiv:2603.12930},
  year         = {2026},
  url          = {https://arxiv.org/abs/2603.12930}
}

@ARTICLE{Mayer2019ForSim,
  author={Mayer, Owen and Stamm, Matthew C.},
  journal={IEEE Transactions on Information Forensics and Security}, 
  title={{Forensic Similarity for Digital Images}}, 
  year={2020},
  volume={15},
  number={},
  pages={1331-1346},
  doi={10.1109/TIFS.2019.2924552}}

@ARTICLE{DSO,
  author={de Carvalho, Tiago José and Riess, Christian and Angelopoulou, Elli and Pedrini, Hélio and de Rezende Rocha, Anderson},
  journal={IEEE Transactions on Information Forensics and Security}, 
  title={{Exposing Digital Image Forgeries by Illumination Color Classification}}, 
  year={2013},
  volume={8},
  number={7},
  pages={1182-1194},
  doi={10.1109/TIFS.2013.2265677}}

@INPROCEEDINGS{Columbia,
  author={Hsu, Yu-feng and Chang, Shih-fu},
  booktitle={2006 IEEE International Conference on Multimedia and Expo}, 
  title={{Detecting Image Splicing using Geometry Invariants and Camera Characteristics Consistency}}, 
  year={2006},
  volume={},
  number={},
  pages={549-552},
  doi={10.1109/ICME.2006.262447}}

@misc {NIST,
  title={{NIST Nimble 2016 Datasets}},
  author={{National Institute of Standards and Technology (NIST)}},
  howpublished={\url{https://www.nist.gov/itl/iad/mig/nimble-challenge-2017-evaluation/}},
  year={2016}
}

@INPROCEEDINGS{CASIA,
  author={Dong, Jing and Wang, Wei and Tan, Tieniu},
  booktitle={2013 IEEE China Summit and International Conference on Signal and Information Processing}, 
  title={{CASIA Image Tampering Detection Evaluation Database}}, 
  year={2013},
  volume={},
  number={},
  pages={422-426},
  doi={10.1109/ChinaSIP.2013.6625374}}

@ARTICLE{Antonutti2026FaceMA,
  author={Antonutti, Davide and Ilyes, Justin and Ruzicka, Laurenz and Poletti, Silvia and Hasenbalg, Marcel and Boyer, Martin and Fischinger, David},
  journal={IEEE Transactions on Biometrics, Behavior, and Identity Science}, 
  title={{Face Morphing Attack Generation and Detection: A State-of-the-Art Review}},
  year={2026},
  volume={8},
  number={3},
  pages={412-430},
  doi={10.1109/TBIOM.2026.3655515}}

@inproceedings{soleymani_differential_2021,
	title = {Differential {Morphed} {Face} {Detection} {Using} {Deep} {Siamese} {Networks}},
	isbn = {978-3-030-68780-9},
	booktitle = {Pattern {Recognition}. {ICPR} {International} {Workshops} and {Challenges}},
	publisher = {Springer International Publishing},
    address = {Cham},
	author = {Soleymani, Sobhan and Chaudhary, Baaria and Dabouei, Ali and Dawson, Jeremy and Nasrabadi, Nasser M.},
	editor = {Del Bimbo, Alberto and Cucchiara, Rita and Sclaroff, Stan and Farinella, Giovanni Maria and Mei, Tao and Bertini, Marco and Escalante, Hugo Jair and Vezzani, Roberto},
	year = {2021},
	pages = {560--572},
}

@INPROCEEDINGS{ferrara_magic_2014,
  author={Ferrara, Matteo and Franco, Annalisa and Maltoni, Davide},
  booktitle={IEEE International Joint Conference on Biometrics}, 
  title = {{The Magic Passport}},
  year={2014},
  volume={},
  number={},
  pages={1-7},
  doi={10.1109/BTAS.2014.6996240}}

@INPROCEEDINGS{ambikapathy_biometric_2024,
  author={Ambikapathy, A. and Beeta, Thiyam Deepa and Kanna, R. Kishore and Danquah-Amoah, Archibald and Ramya, V. Subha and Mutheeswaran, U.},
  booktitle={2024 IEEE International Conference on Computing, Power and Communication Technologies (IC2PCT)}, 
  title={{Biometric Application on Facial Image Recognition Techniques}}, 
  year={2024},
  volume={5},
  number={},
  pages={848-851},
  doi={10.1109/IC2PCT60090.2024.10486608}}

@article{tapia_single-morphing_2025,
title = {{Single-Morphing Attack Detection Using Few-Shot Learning and Triplet-Loss}},
journal = {Neurocomputing},
volume = {636},
pages = {130033},
year = {2025},
issn = {0925-2312},
doi = {https://doi.org/10.1016/j.neucom.2025.130033},
url = {https://www.sciencedirect.com/science/article/pii/S0925231225007052},
author = {Juan E. Tapia and Daniel Schulz and Christoph Busch}
}

@ARTICLE{KAS24,
  author={Khan, Fazlullah and Alturki, Ryan and Srivastava, Gautam and Gazzawe, Foziah and Shah, Syed Tauhid Ullah and Mastorakis, Spyridon},
  journal={IEEE Transactions on Computational Social Systems}, 
  title={{Explainable Detection of Fake News on Social Media Using Pyramidal Co-Attention Network}}, 
  year={2024},
  volume={11},
  number={4},
  pages={4574-4583},
  doi={10.1109/TCSS.2022.3207993}}

@article{LKV22,
author={Orestis Lampridis and Dimitra Karanatsiou and Athena Vakali },
title={{MANIFESTO: a huMAN-centric explaInable approach for FakE news spreaders deTectiOn}},
journal={Computing},
volume={104},
pages={717-739},
year={2022},
doi={10.1007/s00607-021-01013-w}
}

@inproceedings{XLM-R,
    title = "{Unsupervised Cross-lingual Representation Learning at Scale}",
    author = "Conneau, Alexis  and
      Khandelwal, Kartikay  and
      Goyal, Naman  and
      Chaudhary, Vishrav  and
      Wenzek, Guillaume  and
      Guzm{\'a}n, Francisco  and
      Grave, Edouard  and
      Ott, Myle  and
      Zettlemoyer, Luke  and
      Stoyanov, Veselin",
    editor = "Jurafsky, Dan  and
      Chai, Joyce  and
      Schluter, Natalie  and
      Tetreault, Joel",
    booktitle = "Proceedings of the 58th Annual Meeting of the Association for Computational Linguistics",
    month = jul,
    year = "2020",
    address = "Online",
    publisher = "Association for Computational Linguistics",
    url = "https://aclanthology.org/2020.acl-main.747/",
    doi = "10.18653/v1/2020.acl-main.747",
    pages = "8440--8451"
}

@article{MYP21, title={{Machine Learning Explanations to Prevent Overtrust in Fake News Detection}}, volume={15}, url={https://ojs.aaai.org/index.php/ICWSM/article/view/18072}, DOI={10.1609/icwsm.v15i1.18072}, abstractNote={Combating fake news and misinformation propagation is a challenging task in the post-truth era. News feed and search algorithms could potentially lead to unintentional large-scale propagation of false and fabricated information with users being exposed to algorithmically selected false content. Our research investigates the effects of an Explainable AI assistant embedded in news review platforms for combating the propagation of fake news. We design a news reviewing and sharing interface, create a dataset of news stories, and train four interpretable fake news detection algorithms to study the effects of algorithmic transparency on end-users. We present evaluation results and analysis from multiple controlled crowdsourced studies. For a deeper understanding of Explainable AI systems, we discuss interactions between user engagement, mental model, trust, and performance measures in the process of explaining. The study results indicate that explanations helped participants to build appropriate mental models of the intelligent assistants in different conditions and adjust their trust according to their perceptions of model limitations.}, number={1}, journal={Proceedings of the International AAAI Conference on Web and Social Media}, author={Mohseni, Sina and Yang, Fan and Pentyala, Shiva and Du, Mengnan and Liu, Yi and Lupfer, Nic and Hu, Xia and Ji, Shuiwang and Ragan, Eric}, year={2021}, month={May}, pages={421-431} }

@inproceedings{SCW19,
author = {Shu, Kai and Cui, Limeng and Wang, Suhang and Lee, Dongwon and Liu, Huan},
title = {{dEFEND: Explainable Fake News Detection}},
year = {2019},
organizer = {Association for Computing Machinery},
url = {https://doi.org/10.1145/3292500.3330935},
booktitle = {Proceedings of the 25th ACM SIGKDD International Conference on Knowledge Discovery \& Data Mining},
pages = {395–405},
numpages = {11},
series = {KDD '19}
}

@INPROCEEDINGS{ivanovska_selfmad_2025,
  author={Ivanovska, Marija and Todorov, Leon and Damer, Naser and Jain, Deepak Kumar and Peer, Peter and Štruc, Vitomir},
  booktitle={2025 IEEE 19th International Conference on Automatic Face and Gesture Recognition (FG)}, 
  title={{SelfMAD: Enhancing Generalization and Robustness in Morphing Attack Detection via Self-Supervised Learning}}, 
  year={2025},
  volume={},
  number={},
  pages={1-10},
  doi={10.1109/FG61629.2025.11099381}}

@InProceedings{effvit,
author="Coccomini, Davide Alessandro
and Messina, Nicola
and Gennaro, Claudio
and Falchi, Fabrizio",
editor="Sclaroff, Stan
and Distante, Cosimo
and Leo, Marco
and Farinella, Giovanni M.
and Tombari, Federico",
title={{Combining EfficientNet and Vision Transformers for Video Deepfake Detection}},
booktitle="Image Analysis and Processing -- ICIAP 2022",
year="2022",
publisher="Springer International Publishing",
address = {Cham},
pages="219--229",
isbn="978-3-031-06433-3"
}

@article{wodajo2023deepfake,
author = {Wodajo, Deressa and Atnafu, Solomon and Akhtar, Zahid},
year = {2023},
month = {07},
pages = {},
title = {{Deepfake Video Detection Using Generative Convolutional Vision Transformer}},
doi = {10.48550/arXiv.2307.07036}
}

@misc{seferbekov_selimsefdfdc_deepfake_challenge_2026,
	title = {selimsef/dfdc\_deepfake\_challenge},
	copyright = {MIT},
	url = {https://github.com/selimsef/dfdc_deepfake_challenge},
	urldate = {2026-05-06},
	author = {Seferbekov, Selim},
	month = may,
	year = {2026},
	note = {original-date: 2020-06-06T15:22:31Z},
}

@INPROCEEDINGS{dong_02_2023,
  author={Dong, Shichao and Wang, Jin and Ji, Renhe and Liang, Jiajun and Fan, Haoqiang and Ge, Zheng},
  booktitle={2023 IEEE/CVF Conference on Computer Vision and Pattern Recognition (CVPR)}, 
  title={{Implicit Identity Leakage: The Stumbling Block to Improving Deepfake Detection Generalization}}, 
  year={2023},
  volume={},
  number={},
  pages={3994-4004},
  doi={10.1109/CVPR52729.2023.00389}}

@inproceedings{yan2023deepfakebench,
title={{DeepfakeBench: A Comprehensive Benchmark of Deepfake Detection}},
author={Zhiyuan Yan and Yong Zhang and Xinhang Yuan and Siwei Lyu and Baoyuan Wu},
booktitle={Thirty-seventh Conference on Neural Information Processing Systems Datasets and Benchmarks Track},
year={2023},
url={https://openreview.net/forum?id=hizSx8pf0U}
}

@inproceedings{nguyen_capsule-forensics_2019,
	title = {Capsule-forensics: {Using} {Capsule} {Networks} to {Detect} {Forged} {Images} and {Videos}},
	issn = {2379-190X},
	shorttitle = {Capsule-forensics},
	doi = {10.1109/ICASSP.2019.8682602},
	urldate = {2025-01-13},
	booktitle = {{ICASSP} 2019 - 2019 {IEEE} {International} {Conference} on {Acoustics}, {Speech} and {Signal} {Processing} ({ICASSP})},
	author = {Nguyen, Huy H. and Yamagishi, Junichi and Echizen, Isao},
	month = may,
	year = {2019},
	pages = {2307--2311},
}

@INPROCEEDINGS{ni_core_2022,
  author={Ni, Yunsheng and Meng, Depu and Yu, Changqian and Quan, Chengbin and Ren, Dongchun and Zhao, Youjian},
  booktitle={2022 IEEE/CVF Conference on Computer Vision and Pattern Recognition Workshops (CVPRW)}, 
  title={{CORE: Consistent Representation Learning for Face Forgery Detection}}, 
  year={2022},
  volume={},
  number={},
  pages={12-21},
  doi={10.1109/CVPRW56347.2022.00011}}

@InProceedings{tan_efficientnet_2020,
  title = 	 {{EfficientNet: Rethinking Model Scaling for Convolutional Neural Networks}},
  author =       {Tan, Mingxing and Le, Quoc},
  booktitle = 	 {Proceedings of the 36th International Conference on Machine Learning},
  pages = 	 {6105--6114},
  year = 	 {2019},
  editor = 	 {Chaudhuri, Kamalika and Salakhutdinov, Ruslan},
  volume = 	 {97},
  series = 	 {Proceedings of Machine Learning Research},
  month = 	 {09--15 Jun},
  publisher =    {PMLR},
  address = {Long Beach, California, USA},
  url = 	 {https://proceedings.mlr.press/v97/tan19a.html}
}

@inproceedings{qian_thinking_2020,
author = {Qian, Yuyang and Yin, Guojun and Sheng, Lu and Chen, Zixuan and Shao, Jing},
title = {{Thinking in Frequency: Face Forgery Detection by Mining Frequency-Aware Clues}},
year = {2020},
isbn = {978-3-030-58609-6},
publisher = {Springer-Verlag},
address = {Berlin, Heidelberg},
doi = {10.1007/978-3-030-58610-2_6},
booktitle = {Computer Vision – ECCV 2020: 16th European Conference, Glasgow, UK, August 23–28, 2020, Proceedings, Part XII},
pages = {86–103},
numpages = {18},
location = {Glasgow, United Kingdom}
}

@INPROCEEDINGS{afchar_mesonet_2018,
  author={Afchar, Darius and Nozick, Vincent and Yamagishi, Junichi and Echizen, Isao},
  booktitle={2018 IEEE International Workshop on Information Forensics and Security (WIFS)}, 
  title={{MesoNet: a Compact Facial Video Forgery Detection Network}}, 
  year={2018},
  volume={},
  number={},
  pages={1-7},
  doi={10.1109/WIFS.2018.8630761}}

@INPROCEEDINGS{dang_detection_2020,
  author={Dang, Hao and Liu, Feng and Stehouwer, Joel and Liu, Xiaoming and Jain, Anil K.},
  booktitle={2020 IEEE/CVF Conference on Computer Vision and Pattern Recognition (CVPR)}, 
  title={{On the Detection of Digital Face Manipulation}}, 
  year={2020},
  volume={},
  number={},
  pages={5780-5789},
  doi={10.1109/CVPR42600.2020.00582}}

@INPROCEEDINGS{cao_end--end_2022,
  author={Cao, Junyi and Ma, Chao and Yao, Taiping and Chen, Shen and Ding, Shouhong and Yang, Xiaokang},
  booktitle={2022 IEEE/CVF Conference on Computer Vision and Pattern Recognition (CVPR)}, 
  title={{End-to-End Reconstruction-Classification Learning for Face Forgery Detection}}, 
  year={2022},
  volume={},
  number={},
  pages={4103-4112},
  doi={10.1109/CVPR52688.2022.00408}}

@INPROCEEDINGS{liu_spatial-phase_2021,
  author={Liu, Honggu and Li, Xiaodan and Zhou, Wenbo and Chen, Yuefeng and He, Yuan and Xue, Hui and Zhang, Weiming and Yu, Nenghai},
  booktitle={2021 IEEE/CVF Conference on Computer Vision and Pattern Recognition (CVPR)}, 
  title={{Spatial-Phase Shallow Learning: Rethinking Face Forgery Detection in Frequency Domain}}, 
  year={2021},
  volume={},
  number={},
  pages={772-781},
  doi={10.1109/CVPR46437.2021.00083}}

@inproceedings{luo_generalizing_2021,
	title = {Generalizing {Face} {Forgery} {Detection} with {High}-frequency {Features}},
	doi = {10.1109/CVPR46437.2021.01605},
	urldate = {2025-01-13},
	booktitle = {2021 {IEEE}/{CVF} {Conference} on {Computer} {Vision} and {Pattern} {Recognition} ({CVPR})},
	author = {Luo, Yuchen and Zhang, Yong and Yan, Junchi and Liu, Wei},
	month = jun,
	year = {2021},
	pages = {16312--16321},
}

@INPROCEEDINGS{yan_ucf_2023,
  author={Yan, Zhiyuan and Zhang, Yong and Fan, Yanbo and Wu, Baoyuan},
  booktitle={2023 IEEE/CVF International Conference on Computer Vision (ICCV)}, 
  title={{UCF: Uncovering Common Features for Generalizable Deepfake Detection}}, 
  year={2023},
  volume={},
  number={},
  pages={22355-22366},
  doi={10.1109/ICCV51070.2023.02048}}

@INPROCEEDINGS{rossler_faceforensics_2019,
  author={Rössler, Andreas and Cozzolino, Davide and Verdoliva, Luisa and Riess, Christian and Thies, Justus and Niessner, Matthias},
  booktitle={2019 IEEE/CVF International Conference on Computer Vision (ICCV)}, 
  title={{FaceForensics++: Learning to Detect Manipulated Facial Images}}, 
  year={2019},
  volume={},
  number={},
  pages={1-11},
  doi={10.1109/ICCV.2019.00009}}

@INPROCEEDINGS{sun_improving_2021,
  author={Sun, Zekun and Han, Yujie and Hua, Zeyu and Ruan, Na and Jia, Weijia},
  booktitle={2021 IEEE/CVF Conference on Computer Vision and Pattern Recognition (CVPR)}, 
  title={{Improving the Efficiency and Robustness of Deepfakes Detection through Precise Geometric Features}}, 
  year={2021},
  volume={},
  number={},
  pages={3608-3617},
  doi={10.1109/CVPR46437.2021.00361}}

@inproceedings{
yan_orthogonal_2025,
title={{Orthogonal Subspace Decomposition for Generalizable {AI}-Generated Image Detection}},
author={Zhiyuan Yan and Jiangming Wang and Peng Jin and Ke-Yue Zhang and Chengchun Liu and Shen Chen and Taiping Yao and Shouhong Ding and Baoyuan Wu and Li Yuan},
booktitle={Forty-second International Conference on Machine Learning},
year={2025},
url={https://openreview.net/forum?id=GFpjO8S8Po}
}

@misc{midjourney5,
  title        = {{Midjourney 5.2}},
  author       = {{Midjourney, Inc.}},
  year         = {2024},
  howpublished = {Online},
  url          = {https://docs.midjourney.com/docs/model-version-5},
  note         = {Accessed: 2025-01-16}
}

@misc{midjourney6,
  title        = {{Midjourney 6.1}},
  author       = {{Midjourney, Inc.}},
  year         = {2024},
  howpublished = {Online},
  url          = {https://docs.midjourney.com/docs/models},
  note         = {Accessed: 2025-01-16}
}

@InProceedings{glide,
  title = 	 {{{GLIDE}: Towards Photorealistic Image Generation and Editing with Text-Guided Diffusion Models}},
  author =       {Nichol, Alexander Quinn and Dhariwal, Prafulla and Ramesh, Aditya and Shyam, Pranav and Mishkin, Pamela and Mcgrew, Bob and Sutskever, Ilya and Chen, Mark},
  booktitle = 	 {Proceedings of the 39th International Conference on Machine Learning},
  pages = 	 {16784--16804},
  year = 	 {2022},
  editor = 	 {Chaudhuri, Kamalika and Jegelka, Stefanie and Song, Le and Szepesvari, Csaba and Niu, Gang and Sabato, Sivan},
  volume = 	 {162},
  series = 	 {Proceedings of Machine Learning Research},
  month = 	 {17--23 Jul},
  publisher =    {PMLR},
  address = {Baltimore, Maryland, USA},
  url = 	 {https://proceedings.mlr.press/v162/nichol22a.html}
}

@article{sdxl,
	title={SDXL: Improving Latent Diffusion Models for High-Resolution Image Synthesis},
	author={Podell, Dustin and English, Zion and Lacey, Kyle and Blattmann, Andreas and Dockhorn, Tim and Müller, Jonas and Penna, Joe and Rombach, Robin},
	journal={arXiv preprint arXiv:2307.01952},
	year={2023},
	url={https://arxiv.org/abs/2307.01952}
}

@misc{dalle2,
	author = {OpenAI},
	title = {{DALL-E 2: Generating images from text}},
	year = {2022},
	howpublished = {\url{https://openai.com/index/dall-e-2}},
	note = {Accessed: 2025-01-16}
}

@misc{dalle3,
	author = {OpenAI},
	title = {{DALL-E 3: Generating images from text}},
	year = {2023},
	howpublished = {\url{https://openai.com/index/dall-e-3}},
	note = {Accessed: 2025-01-16}
}

@article{openimg,
	title={{The Open Images Dataset V4: Unified Image Classification, Object Detection, and Visual Relationship Detection at Scale}},
	volume={128},
	ISSN={1573-1405},
	url={http://dx.doi.org/10.1007/s11263-020-01316-z},
	DOI={10.1007/s11263-020-01316-z},
	number={7},
	journal={International Journal of Computer Vision},
	publisher={Springer Science and Business Media LLC},
	author={Kuznetsova, Alina and Rom, Hassan and Alldrin, Neil and Uijlings, Jasper and Krasin, Ivan and Pont-Tuset, Jordi and Kamali, Shahab and Popov, Stefan and Malloci, Matteo and Kolesnikov, Alexander and Duerig, Tom and Ferrari, Vittorio},
	year={2020},
	month=mar, pages={1956–1981} }

@misc{pascal-voc-2012,
	author = "Everingham, M. and Van~Gool, L. and Williams, C. K. I. and Winn, J. and Zisserman, A.",
	title = "The {PASCAL} {V}isual {O}bject {C}lasses {C}hallenge 2012 {(VOC2012)} {R}esults",
	year={2015},
	howpublished = {http://www.pascal-network.org/challenges/VOC/voc2012/workshop/index.html}
}

@misc{flux,
  title        = {{Flux}},
  author       = {{Black Forest Labs}},
  year         = {2024},
  howpublished = {Online},
  url          = {https://github.com/black-forest-labs/flux},
  note         = {Accessed: 2026-06-19}
}

@misc{firefly,
	author = {Adobe},
	title = {{Adobe Firefly Text To Image}},
	year = {2024},
	howpublished = {\url{https://firefly.adobe.com}},
	note = {Accessed: 2025-01-16}
}

@INPROCEEDINGS{imagenet,
	author={Deng, Jia and Dong, Wei and Socher, Richard and Li, Li-Jia and Kai Li and Li Fei-Fei},
	booktitle={2009 IEEE Conference on Computer Vision and Pattern Recognition}, 
	title={{ImageNet: A Large-Scale Hierarchical Image Database}}, 
	year={2009},
	volume={},
	number={},
	pages={248-255},
	doi={10.1109/CVPR.2009.5206848}
}

@INPROCEEDINGS{yu_006049_2021,
  author={Yu, Ning and Skripniuk, Vladislav and Abdelnabi, Sahar and Fritz, Mario},
  booktitle={2021 IEEE/CVF International Conference on Computer Vision (ICCV)}, 
  title={{Artificial Fingerprinting for Generative Models: Rooting Deepfake Attribution in Training Data}}, 
  year={2021},
  volume={},
  number={},
  pages={14428-14437},
  doi={10.1109/ICCV48922.2021.01418}}

@misc{noauthor_010000_2024,
  title        = {{Identifying AI-generated Images}},
  author       = {{SynthID}},
  year         = {2026},
  howpublished = {Online},
  url          = {https://deepmind.google/discover/blog/identifying-ai-generated-images-with-synthid/},
  note         = {Accessed: 2024-11-06}
}

@article{borji2023qualitative,
title = {{Qualitative Failures of Image Generation Models and Their Application in Detecting Deepfakes}},
journal = {Image and Vision Computing},
volume = {137},
pages = {104771},
year = {2023},
issn = {0262-8856},
doi = {https://doi.org/10.1016/j.imavis.2023.104771},
url = {https://www.sciencedirect.com/science/article/pii/S0262885623001452},
author = {Ali Borji}
}

@misc{zhong_007101_2024,
	title = {{PatchCraft}: {Exploring} {Texture} {Patch} for {Efficient} {AI}-generated {Image} {Detection}}, 
	booktitle= {{PatchCraft}: {Exploring} {Texture} {Patch} for {Efficient} {AI}-generated {Image} {Detection}},
	shorttitle = {{PatchCraft}},
	url = {http://arxiv.org/abs/2311.12397},
	author = {Zhong, Nan and Xu, Yiran and Li, Sheng and Qian, Zhenxing and Zhang, Xinpeng},
	month = mar,
	year = {2024},
}

@InProceedings{frank_007902_2020,
  title = 	 {{Leveraging Frequency Analysis for Deep Fake Image Recognition}},
  author =       {Frank, Joel and Eisenhofer, Thorsten and Sch{\"o}nherr, Lea and Fischer, Asja and Kolossa, Dorothea and Holz, Thorsten},
  booktitle = 	 {Proceedings of the 37th International Conference on Machine Learning},
  pages = 	 {3247--3258},
  year = 	 {2020},
  editor = 	 {III, Hal Daumé and Singh, Aarti},
  volume = 	 {119},
  series = 	 {Proceedings of Machine Learning Research},
  month = 	 {13--18 Jul},
  publisher =    {PMLR},
  address = {Virtual},
  url = 	 {https://proceedings.mlr.press/v119/frank20a.html},
}

@INPROCEEDINGS{corvi_007160_2023,
  author={Corvi, Riccardo and Cozzolino, Davide and Poggi, Giovanni and Nagano, Koki and Verdoliva, Luisa},
  booktitle={2023 IEEE/CVF Conference on Computer Vision and Pattern Recognition Workshops (CVPRW)}, 
  title={{Intriguing properties of synthetic images: from generative adversarial networks to diffusion models}}, 
  year={2023},
  volume={},
  number={},
  pages={973-982},
  doi={10.1109/CVPRW59228.2023.00104}}

@INPROCEEDINGS{gradcam,
  author={Selvaraju, Ramprasaath R. and Cogswell, Michael and Das, Abhishek and Vedantam, Ramakrishna and Parikh, Devi and Batra, Dhruv},
  booktitle={2017 IEEE International Conference on Computer Vision (ICCV)}, 
  title={{Grad-CAM: Visual Explanations from Deep Networks via Gradient-Based Localization}}, 
  year={2017},
  volume={},
  number={},
  pages={618-626},
  doi={10.1109/ICCV.2017.74}}

@inproceedings{shap,
author = {Lundberg, Scott M. and Lee, Su-In},
title = {{A Unified Approach to Interpreting Model Predictions}},
year = {2017},
isbn = {9781510860964},
publisher = {Curran Associates Inc.},
address = {Red Hook, NY, USA},
booktitle = {Proceedings of the 31st International Conference on Neural Information Processing Systems},
pages = {4768–4777},
numpages = {10},
location = {Long Beach, California, USA},
series = {NIPS'17}
}

@misc{poletti2026aigeneratedimageshumansmachines,
      title={{AI-Generated Images: What Humans and Machines See When They Look at the Same Image}}, 
      author={Silvia Poletti and Justin Ilyes and Marcel Hasenbalg and David Fischinger and Martin Boyer},
      year={2026},
      eprint={2605.06143},
      archivePrefix={arXiv},
      primaryClass={cs.CV},
      url={https://arxiv.org/abs/2605.06143}, 
}

@INPROCEEDINGS{resnet,
  author={He, Kaiming and Zhang, Xiangyu and Ren, Shaoqing and Sun, Jian},
  booktitle={2016 IEEE Conference on Computer Vision and Pattern Recognition (CVPR)}, 
  title={{Deep Residual Learning for Image Recognition}}, 
  year={2016},
  volume={},
  number={},
  pages={770-778},
  doi={10.1109/CVPR.2016.90}}

@inproceedings{vit,
title={{An Image is Worth 16x16 Words: Transformers for Image Recognition at Scale}},
author={Alexey Dosovitskiy and Lucas Beyer and Alexander Kolesnikov and Dirk Weissenborn and Xiaohua Zhai and Thomas Unterthiner and Mostafa Dehghani and Matthias Minderer and Georg Heigold and Sylvain Gelly and Jakob Uszkoreit and Neil Houlsby},
booktitle={International Conference on Learning Representations},
year={2021},
url={https://openreview.net/forum?id=YicbFdNTTy}
}

@inproceedings{zhu_001000_2024,
 author = {Zhu, Mingjian and Chen, Hanting and Yan, Qiangyu and Huang, Xudong and Lin, Guanyu and Li, Wei and Tu, Zhijun and Hu, Hailin and Hu, Jie and Wang, Yunhe},
 booktitle = {Advances in Neural Information Processing Systems},
 doi = {10.52202/075280-3398},
 editor = {A. Oh and T. Naumann and A. Globerson and K. Saenko and M. Hardt and S. Levine},
 pages = {77771--77782},
 publisher = {Curran Associates, Inc.},
 address = {New Orleans, Louisiana, USA},
 title = {{GenImage: A Million-Scale Benchmark for Detecting AI-Generated Image}},
 volume = {36},
 year = {2023}
}

@inproceedings{dogoulis2023improving,
author = {Dogoulis, Pantelis and Kordopatis-Zilos, Giorgos and Kompatsiaris, Ioannis and Papadopoulos, Symeon},
title = {{Improving Synthetically Generated Image Detection in Cross-Concept Settings}},
year = {2023},
isbn = {9798400701870},
publisher = {Association for Computing Machinery},
address = {New York, NY, USA},
url = {https://doi.org/10.1145/3592572.3592846},
doi = {10.1145/3592572.3592846},
booktitle = {Proceedings of the 2nd ACM International Workshop on Multimedia AI against Disinformation},
pages = {28–35},
numpages = {8},
location = {Thessaloniki, Greece},
series = {MAD '23}
}

@misc{vera2026plugin,
  title        = {{Verification Plugin}},
  author       = {{Vera AI}},
  year         = {2026},
  howpublished = {Online},
  url          = {https://www.veraai.eu/category/verification-plugin},
  note         = {Accessed: 2026-06-01}
}

@misc{milner2026verificationassistant,
  title         = {{A Browser-based Open Source Assistant for Multimodal Content Verification}},
  author        = {Milner, Rosanna and Foster, Michael and Razuvayevskaya, Olesya and Roberts, Ian and Porcellini, Valentin and Teyssou, Denis and Bontcheva, Kalina},
  year          = {2026},
  eprint        = {2603.02842},
  archivePrefix = {arXiv},
  primaryClass  = {cs.CL},
  url           = {https://arxiv.org/abs/2603.02842}
}

@INPROCEEDINGS{ju2024deepfakeometer,
  author={Ju, Yan and Sun, Chengzhe and Jia, Shan and Hou, Shuwei and Si, Zhaofeng and Datta, Soumyya Kanti and Ke, Lipeng and Zhou, Riky and Nikolich, Anita and Lyu, Siwei},
  booktitle={2024 IEEE 7th International Conference on Multimedia Information Processing and Retrieval (MIPR)}, 
  title={{DeepFake-o-meter v2.0: An Open Platform for DeepFake Detection}}, 
  year={2024},
  volume={},
  number={},
  pages={439-445},
  doi={10.1109/MIPR62202.2024.00075}}

@misc{truemedia2026detector,
  title        = {{AI Media Detector}},
  author       = {{TrueMedia}},
  year         = {2026},
  howpublished = {Online},
  url          = {https://truemedia.app/},
  note         = {Accessed: 2026-06-01}
}

@misc{sensity2026platform,
  title        = {{Deepfake Detection for Video, Image \& Audio}},
  author       = {{Sensity AI}},
  year         = {2026},
  howpublished = {Online},
  url          = {https://sensity.ai/deepfake-detection-for-video-image-audio/},
  note         = {Accessed: 2026-06-01}
}

@misc{realitydefender2025sdk,
  title        = {{Reality Defender Brings Deepfake Detection to Developers with New SDK Powered by NVIDIA}},
  author       = {{Reality Defender}},
  year         = {2025},
  howpublished = {Online},
  url          = {https://www.realitydefender.com/insights/reality-defender-brings-deepfake-detection-to-developers-with-new-sdk-powered-by-nvidia},
  note         = {Accessed: 2026-06-01}
}

@article{deng_arcface_2019,
	title = {{ArcFace}: {Additive} {Angular} {Margin} {Loss} for {Deep} {Face} {Recognition}},
	language = {en},
	author = {Deng, Jiankang and Guo, Jia and Xue, Niannan and Zafeiriou, Stefanos},
    year = {2019},
    journal={2019 IEEE/CVF Conference on Computer Vision and Pattern Recognition (CVPR)},
}

@inproceedings{liu_forgery-aware_2024,
	title = {Forgery-aware {Adaptive} {Transformer} for {Generalizable} {Synthetic} {Image} {Detection}},
	issn = {2575-7075},
	url = {https://ieeexplore.ieee.org/document/10657197/},
	doi = {10.1109/CVPR52733.2024.01024},
	urldate = {2026-04-28},
	booktitle = {2024 {IEEE}/{CVF} {Conference} on {Computer} {Vision} and {Pattern} {Recognition} ({CVPR})},
	author = {Liu, Huan and Tan, Zichang and Tan, Chuangchuang and Wei, Yunchao and Wang, Jingdong and Zhao, Yao},
	month = jun,
	year = {2024},
	pages = {10770--10780},
}

@INPROCEEDINGS{fang_unsupervised_2022,
  author={Fang, Meiling and Boutros, Fadi and Damer, Naser},
  booktitle={2022 IEEE International Joint Conference on Biometrics (IJCB)}, 
  title={{Unsupervised Face Morphing Attack Detection via Self-paced Anomaly Detection}}, 
  year={2022},
  volume={},
  number={},
  pages={1-11},
  doi={10.1109/IJCB54206.2022.10008003}}

@article{IVANOVSKA2026103921,
title = {{SelfMAD++: Self-supervised Foundation Model with Local Feature Enhancement for Generalized Morphing Attack Detection}},
journal = {Information Fusion},
volume = {127},
pages = {103921},
year = {2026},
issn = {1566-2535},
doi = {https://doi.org/10.1016/j.inffus.2025.103921},
url = {https://www.sciencedirect.com/science/article/pii/S1566253525009832},
author = {Marija Ivanovska and Leon Todorov and Peter Peer and Vitomir Štruc}
}

@misc{sota_01_xlsr,
      title={{XLSR-MamBo: Scaling the Hybrid Mamba-Attention Backbone for Audio Deepfake Detection}}, 
      author={Kwok-Ho Ng and Tingting Song and Yongdong Wu and Zhihua Xia},
      year={2026},
      eprint={2601.02944},
      archivePrefix={arXiv},
      primaryClass={eess.AS},
      url={https://arxiv.org/abs/2601.02944}, 
}
\end{document}